\documentclass[runningheads]{llncs}
\usepackage{graphicx}
\usepackage{comment}
\usepackage{amsmath,amssymb} 
\usepackage{color}
\usepackage{commath}
\usepackage[nocompress]{cite}
\usepackage{hyperref}
\hypersetup{
    colorlinks=true,
    linkcolor=magenta,
    filecolor=magenta,      
    urlcolor=magenta,
}

\begin{document}
\pagestyle{headings}
\mainmatter

\title{HDNet: Human Depth Estimation for\\Multi-Person Camera-Space Localization} 


\titlerunning{HDNet: Human Depth Estimation}
%
\author{Jiahao Lin \and Gim Hee Lee}
\authorrunning{Jiahao Lin and Gim Hee Lee}
%
\institute{Department of Computer Science, National University of Singapore
\email{\{jiahao,gimhee.lee\}@comp.nus.edu.sg}}
\maketitle

\begin{abstract}
Current works on multi-person 3D pose estimation mainly focus on the estimation of the 3D joint locations relative to the root joint
and ignore the absolute locations of each pose. In this paper, we propose the Human Depth Estimation Network (HDNet), an end-to-end framework for absolute root joint localization in the camera coordinate space. Our HDNet first estimates the 2D human pose with heatmaps of the joints. These estimated heatmaps serve as attention masks for pooling features from image regions corresponding to the target person. A skeleton-based Graph Neural Network (GNN) is utilized to propagate features among joints. We formulate the target depth regression as a bin index estimation problem, which can be transformed with a soft-argmax operation from the classification output of our HDNet. We evaluate our HDNet on the root joint localization and root-relative 3D pose estimation tasks with two benchmark datasets, \textit{i.e.}, Human3.6M and MuPoTS-3D. The experimental results show that we outperform the previous state-of-the-art consistently under multiple evaluation metrics. Our source code is available at: \url{https://github.com/jiahaoLjh/HumanDepth}.
\end{abstract}

\section{Introduction}
Human pose estimation is one of the active research topics in the community of computer vision and artificial intelligence due to its importance in many applications such as camera surveillance, virtual/augmented reality, and human-computer interaction, \textit{etc}.
Extensive research has been done for human pose estimation in both 2D image space and 3D Cartesian space, respectively. Great successes have been achieved in the single-person 2D/3D pose estimation tasks thanks to the rapid development of deep learning techniques and the emergence of large-scale human pose datasets \cite{andriluka14cvpr,lin2014microsoft,ionescu2014human3}. On the other hand,
multi-person 2D/3D pose estimation tasks are more challenging due to the unknown number of persons in the scene.
To mitigate this problem, the multi-person pose estimation task is typically tackled in a two-stage scheme that decouples the estimation of the number of persons and the pose of each person, \textit{i.e.}, the top-down \cite{he2017mask,fang2017rmpe,huang2017coarse,chen2018cascaded,sun2019deep} or bottom-up \cite{newell2017associative,cao2017realtime,papandreou2018personlab} scheme.
In recent years, large-scale multi-person 3D pose datasets such as MuPoTS-3D \cite{singleshotmultiperson2018} are created to faciliate the research of multi-person 3D pose estimation. However, most of the existing works \cite{dabral2019multi,rogez2017lcr,rogez2019lcr,singleshotmultiperson2018} focus on the estimation of 3D pose relative to the root joint of each person in the scene. Global absolute locations of the respective 3D poses with respect to the camera coordinate space are ignored.

\begin{figure}[t]
\centering
\includegraphics[width=0.85\textwidth,trim={10 30 10 30},clip]{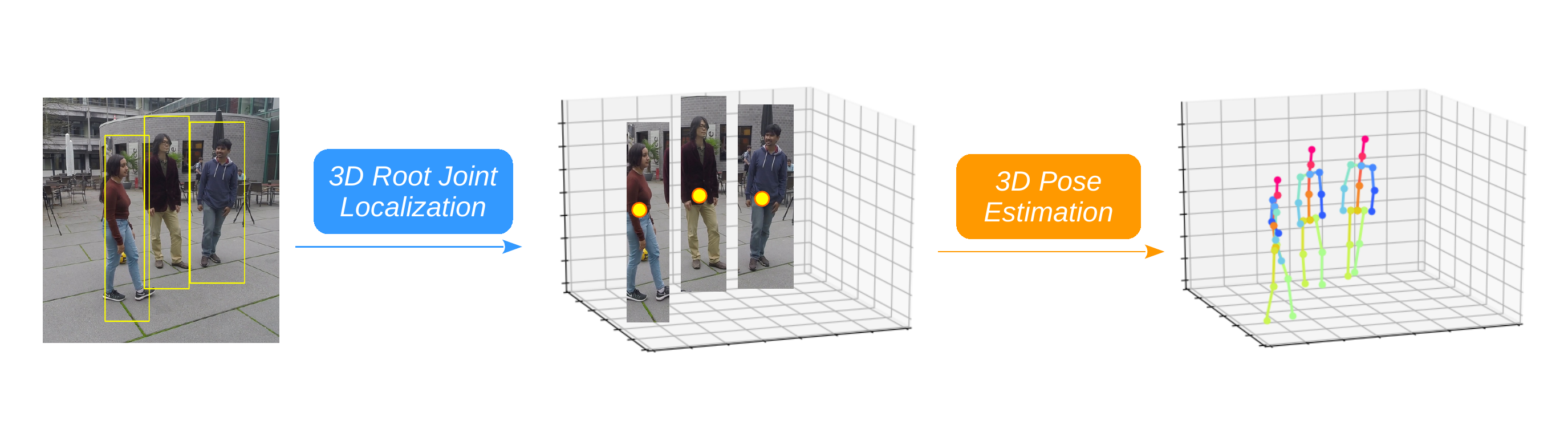}
\caption{Top-down multi-person 3D pose estimation pipeline. Camera-space root joint coordinate is estimated for each detected person bounding box, followed by a root-relative 3D pose estimation, to obtain the absolute 3D poses and locations.}
\label{fig:teaser}
\end{figure}

Estimating the absolute 3D location of each pose in an image is essential for understanding human-to-human interactions.
Recently, Moon \textit{et al.} \cite{Moon_2019_ICCV_3DMPPE} propose a multi-stage pipeline for the task of multi-person 3D pose estimation in the camera coordinate space as shown in Figure \ref{fig:teaser}. The pipeline adopts the top-down scheme which predicts a bounding box for each person in the first stage. 
This is followed by estimation of the absolute root joint location for the person in each bounding box. Finally, the global pose of each person is recovered by applying single-person 3D pose estimation to get the relative location of other joints with respect to the root joint. 
The root joint localization framework proposed in \cite{Moon_2019_ICCV_3DMPPE} estimates the depth of root joint for each person based on the size of the bounding box. Despite showing promising results, the approach relies on the size of bounding box for root joint localization, and hence is not sufficiently effective due to two reasons: (1) The compactness of bounding boxes varies from person to person and also between different object detectors. (2) Sizes of bounding boxes carries no direct information about the size of the particular person due to the variation of poses.

In this paper, we propose an end-to-end Human Depth Estimation Network (HDNet) to address the problems of root joint depth estimation and localization. 
We adopt the same top-down pipeline for the task of multi-person absolute 3D pose estimation.
Our key observation is that we can estimate the depth of a person in a monocular image with considerably high accuracy by leveraging on the prior knowledge of the typical size of the human pose and body joints.
Inspired by this observation, 
we propose to jointly learn the 2D human pose and the depth estimation tasks in our HDNet.
More specifically, we utilize the heatmaps of the joints from the human pose estimation task as attention masks to achieve pose-aware feature pooling in each joint type.
Subsequently, we put the pose-aware features of the joints into a skeleton-based Graph Neural Network (GNN), where information are effectively propagated among body joints to enhance depth estimation.
Following a recent work on scene depth estimation \cite{fu2018deep}, we formulate our depth estimation of the root joint as a classification task, where the target depths are discretized into a preset number of bins.
We also adopt a soft-argmax operation on the bins predicted by our HDNet for faster convergence during training and better performance without losing precision compared to direct numerical depth regression.
Our approach outperforms previous state-of-the-art \cite{Moon_2019_ICCV_3DMPPE} on the task of root joint localization on two benchmark datasets, \textit{i.e.}, Human3.6M \cite{ionescu2014human3} and MuPoTS-3D \cite{singleshotmultiperson2018} under multiple evaluation metrics. 
Experimental results also show that accurate root localization benefits the task of root-aligned 3D human pose estimation.

Our contributions in this work are:
\begin{itemize}
    \item An end-to-end Human Depth Estimation Network (HDNet) is proposed to address the problem of root joint localization for multi-person 3D pose estimation in the camera-space.
    \item Several key components are introduced in our framework: (1) pose heatmaps are used as attention masks for pose-aware feature pooling; (2) a skeleton-based GNN is designed for effective information propagation among the body joints; and (3) 
    depth regression of the root joint is formulated as a classification task, where the classification output is transformed to the estimated depth with a soft-argmax operation to facilitate accurate depth estimation.
    \item Quantitative and qualitative results show that our approach consistently outperforms the state-of-the-art on multiple benchmark datasets under various evaluation metrics.
\end{itemize}

\section{Related Works}

Human pose estimation has been an interesting yet challenging problem in computer vision. Early methods use a variety of hand-crafted features such as silhouette, shape, SIFT features, HOG for the task. Recently, with the power of deep neural networks and well-annotated large-scale human pose datasets, increasing learning-based approaches are proposed to tackle this challenging problem.

\subsubsection{Single-person 2D pose estimation.}
Early works, such as Stacked Hourglass \cite{newell2016stacked}, Convolutional Pose Machines \cite{wei2016convolutional}, \textit{etc.}, have been proposed to use deep convolutional neural networks as feature extractors for 2D pose estimation. Heatmaps of joints are the commonly used representation to indicate the presence of joints at spatial locations with Gaussian peaks. More recent works including RMPE \cite{fang2017rmpe}, CFN \cite{huang2017coarse}, CPN \cite{chen2018cascaded}, HRNet \cite{sun2019deep}, \textit{etc.}, introduce various framework designs to improve the joint localization precision.

\subsubsection{Single-person 3D pose estimation.}
Approaches for 3D pose estimation can be generally categorized into two groups. Direct end-to-end estimation of 3D pose from RGB images regresses both 2D joint locations and the $z$-axis root-relative depth for each joint. \cite{pavlakos2017coarse,sun2018integral} extend the notion of heatmap to the 3D space, where estimation is performed in a volumetric space. Another group of approaches decouples the task into a two-stage pipeline. The 2D joint locations are first estimated, followed by a 2D-to-3D lifting. \cite{martinez2017simple,fang2018learning} utilize Multi-Layer Perceptron (MLP) to learn the mapping.

\subsubsection{Multi-person 2D pose estimation.}
Top-down \cite{he2017mask,fang2017rmpe,huang2017coarse,chen2018cascaded,sun2019deep} and bottom-up \cite{newell2017associative,cao2017realtime,papandreou2018personlab} approaches have been proposed to estimate poses for multiple persons. Top-down approaches utilize a human object detector to localize the bounding box, followed by a single-person pose estimation pipeline with image patch cropped from the bounding box. Bottom-up approaches detect human joints in a person-agnostic way, followed by a grouping process to identify joints belonging to the same person. Top-down approaches usually estimate joint locations more precisely because bounding boxes of different sizes are scaled to the same size in the single-person estimation stage. However, top-down approaches tend to be more computationally expensive due to the redundancy in bounding box detections.

\subsubsection{Multi-person 3D pose estimation.}
Several works \cite{rogez2017lcr,rogez2019lcr,singleshotmultiperson2018,dabral2019multi,zanfir2018deep} have been conducted on multi-person 3D pose estimation.
Rogez \textit{et al.} \cite{rogez2017lcr} propose a LCR-Net which consists of localization, classification, and regression parts and estimates each detected human with a classified and refined anchor pose.
Mehta \textit{et al.} \cite{singleshotmultiperson2018} propose a bottom-up approach which estimates a specially designed occlusion-robust pose map and readout the 3D poses given 2D poses obtained with Part Affinity Fields \cite{cao2017realtime}.
Dabral \textit{et al.} \cite{dabral2019multi} propose to incorporate hourglass network into Mask R-CNN detection heads for better 2D pose localization, followed by a standard residual network to lift 2D poses to 3D.
Zanfir \textit{et al.} \cite{zanfir2018deep} design a holistic multi-person sensing pipeline, \textit{i.e.} MubyNet, to jointly address the problems of multi-person 2D/3D skeleton/shape-based pose estimation.
However, these works only estimate and evaluate the 3D pose after root joint alignment and ignore the global location of each pose.
Recently, Moon \textit{et al.} \cite{Moon_2019_ICCV_3DMPPE} propose a multi-stage pipeline for multi-person camera-space 3D pose estimation. The pipeline follows the top-down scheme and consists of a RootNet which localizes the root joint for each detected bounding box. 
We also adopt the top-down scheme pipeline and estimate the camera-space root joint location and 3D pose for each detected bounding box.
To our best knowledge, \cite{Moon_2019_ICCV_3DMPPE} and our work are the only two works that focus on the estimation and evaluation of multi-person root joint locations.
Compared to \cite{Moon_2019_ICCV_3DMPPE} which relies on the size of detected bounding box, we utilize the underlying features and design a human-specific pose-based root joint depth estimation framework to significantly boost the root localization performance.

\begin{figure}[t]
\centering
\includegraphics[width=0.85\textwidth,trim={10 20 10 10},clip]{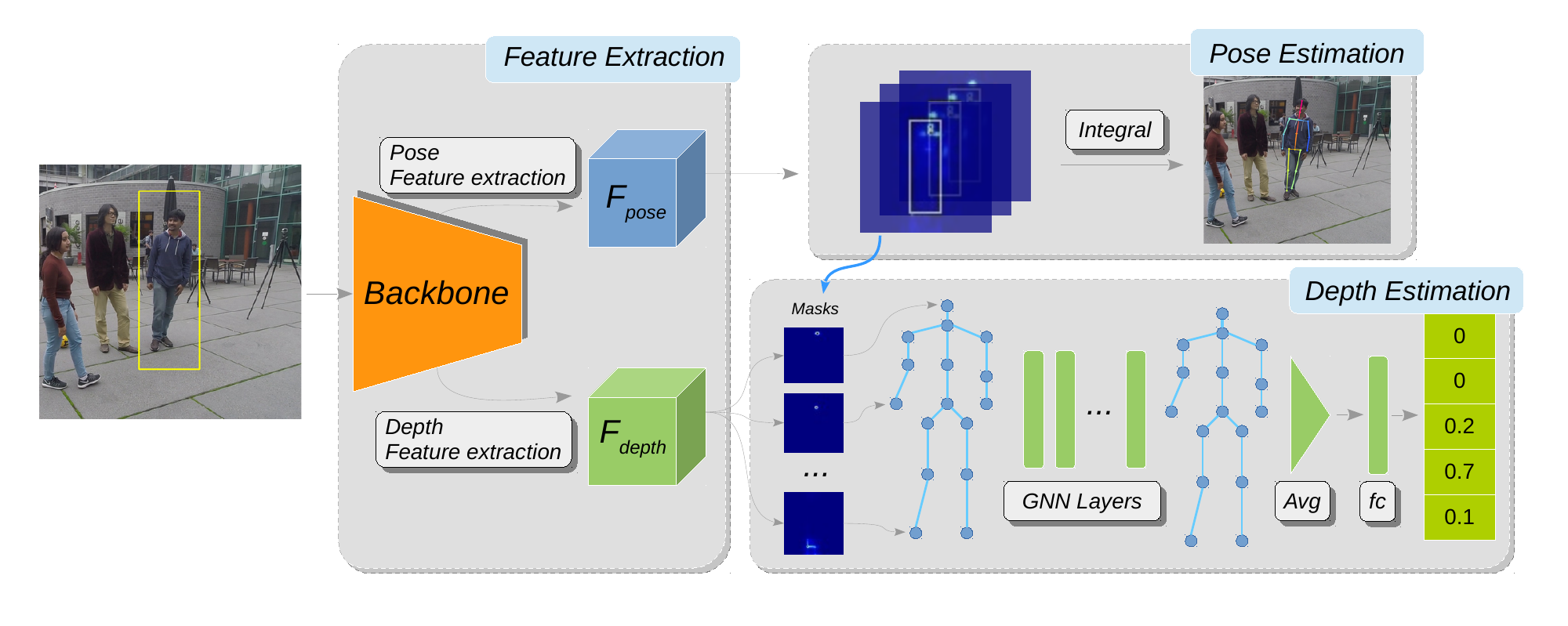}
\caption{Our HDNet architecture. The framework takes an image together with the bounding box of a target person as input. A Feature Pyramid Network backbone is used for general feature extraction followed by separated multi-scale feature extraction for the tasks of pose and depth estimation. Estimated heatmaps are used as attention masks to pool depth features. A Graph Neural Network is utilized to propagate and aggregate features for the target person depth estimation.}
\label{fig:framework}
\end{figure}

\section{Our Approach}

\subsection {Overview}
Given a 2D image with an unknown number of persons, the task of camera-space multi-person 3D pose estimation is to: (1) identify all person instances, (2) estimate the 3D pose with respect to the root joint, \textit{i.e.}, pelvis, for each person, and (3) localize each person by estimating the 3D coordinate of root joint in the camera coordinate space.

Following the top-down approaches in the literature of multi-person pose estimation, we assume that the 2D human bounding boxes for each person in the input image are available from a generic object detector.
Given the person instances and detected bounding boxes, we propose an end-to-end depth estimation framework to localize the root joint of each person in the camera coordinate space as illustrated in Figure \ref{fig:framework}. The root joint localization is decoupled into two sub-tasks: (1) localization of the root joint image coordinate ($u, v$), and (2) estimation of the root joint depth $Z$ in the camera frame, which is then used to back-project ($u, v$) to 3D space.
We use an off-the-shelf single-person 3D pose estimator to estimate the 3D joint locations of each person with respect to the root joint.
The final absolute 3D pose of each person in the camera coordinate system is obtained by the transformation of each joint location with the absolute location of the root joint.

The details of our proposed root joint localization framework are introduced in Section \ref{sec:root_localization_framework}. The choices of specific object detector and single-person 3D pose estimator used in our experiments are given in the implementation details in Section \ref{sec:implementation_details}.

\subsection{Root Localization Framework}
\label{sec:root_localization_framework}

Our framework for monocular image single/multi-person depth estimation is shown in Figure \ref{fig:framework}. The framework consists of a Feature Pyramid Network (FPN)-based backbone, a heatmap-based human pose estimation branch, and a Graph Neural Network (GNN)-based depth estimation branch.

\begin{figure}[t]
\centering
\includegraphics[width=0.85\textwidth,trim={0 10 0 0},clip]{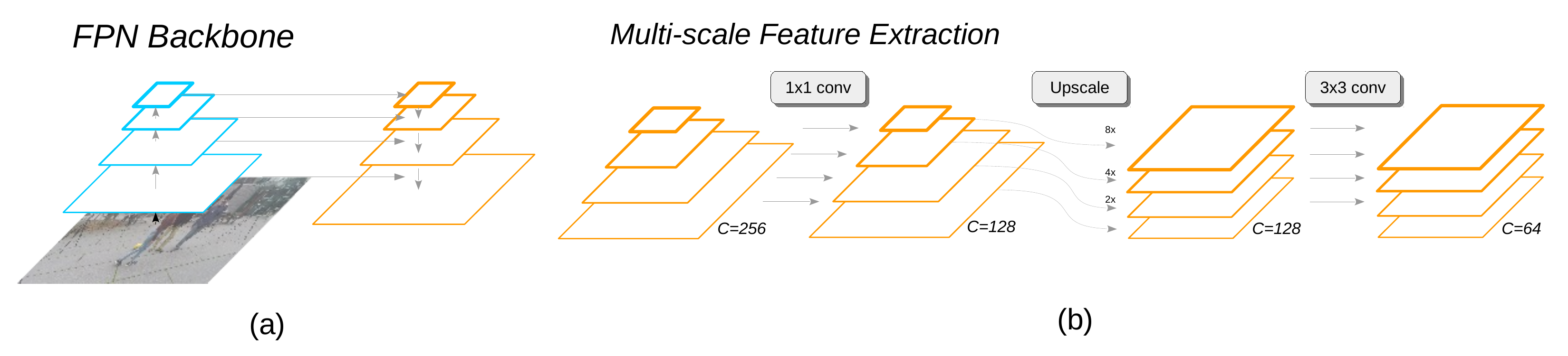}
\caption{(a) ResNet-based Feature Pyramid Network Backbone for general feature extraction. (b) Multi-scale feature extraction subnet architecture used for both Pose feature and Depth feature extraction.}
\label{fig:feature_network}
\end{figure}

\subsubsection{Backbone Network.}
We choose FPN \cite{lin2017feature} as our backbone network due to its capability of explicitly handling features of multiple scales in the form of feature pyramids. Hence, it is suitable for perceiving the scale of human body parts and consequently enhances depth estimation of the human pose in an image. The FPN network consists of a ResNet-50 \cite{he2016deep} with feature blocks of four different scales $C_2, C_3, C_4, C_5$ (cyan layers in Figure \ref{fig:feature_network}(a)), where a reversed hierarchy of feature pyramid $P_5, P_4, P_3, P_2$ is built upon (orange layers in Figure \ref{fig:feature_network}(a)). Each of the four scales encodes hierarchical levels of feature representations, which are then passed through two consecutive convolutional layers as shown in Figure \ref{fig:feature_network}(b). An upsampling operation with corresponding upsample scale factor is applied between the two convolutional layers to ensure matching spatial resolution from the output of the four scales. 
Batch Normalization \cite{ioffe2015batch} and ReLU operations are used after each convolution layer. Weights are not shared across scales. Blocks of all scales are then concatenated to form the final feature block $\mathbf{F}$.
Since we find that the downstream tasks of pose estimation and depth estimation are not collaboratively correlated, we split the multi-scale feature processing from the output feature pyramid $P_5, P_4, P_3, P_2$ of the backbone into two parallel branches without shared weights as shown in Figure \ref{fig:framework}. We denote the features as $\mathbf{F}_{\text{pose}}$ and $\mathbf{F}_{\text{depth}}$, respectively.

\subsubsection{2D Pose Estimation Branch.}
We propose to use estimated 2D pose as a guide to aggregate information from useful feature regions to effectively distil information from the image and discard irrelevant areas such as the background. 
We first regress $N_J$ heatmaps $\hat{\mathbf{H}}$ that correspond to the $N_J$ joints with a $1\times 1$ convolution from feature block $\mathbf{F}_\text{pose}$. Each of the $N_J$ heatmaps are normalized across all spatial locations with a softmax operation. A direct read out of the coordinate from the local maximum limits the precision of the joint location estimation due to the low resolution of the output heatmap (4x downsample from input image in ResNet backbone). To circumvent this problem, we follow the idea of ``soft-argmax" in \cite{sun2018integral} and compute the ``integral" version of estimated coordinate ($\hat{u},\hat{v}$) for each joint $j$ using the weighted sum of coordinates:
\begin{align}
    (\hat{u}_j, \hat{v}_j) = \sum_{(u,v)=(0,0)}^{(W-1,H-1)}\hat{\mathbf{H}}_{u,v}^{(j)} \cdot (u, v),
\label{eqn:integral}
\end{align}
where $W$ and $H$ are the width and height of output heatmap. The softmax operation guarantees that the weights $\hat{\mathbf{H}}_{u,v}$ form a valid distribution which sum up to 1 over all spatial locations.
To supervise the heatmap regression, we generate a
ground truth heatmap $\mathbf{H}^{(j)\text{GT}}$ for each joint $j$. A Gaussian peak is created around the ground truth joint location ($u_j, v_j$) with a preset standard deviation that controls the compactness of the Gaussian peak. We use standard Mean Squared Error (MSE) as the heatmap regression loss and $L1$ loss for the pose after soft-argmax as follows:
\begin{align}
    \mathcal{L}_{\text{hm}}=\frac{1}{N_{J}HW}\sum_j^{N_J}\sum_{(u,v)=(0,0)}^{(W-1,H-1)}\norm{\mathbf{H}_{u,v}^{(j)\text{GT}}-\hat{\mathbf{H}}_{u,v}^{(j)}}^2,\\
    \mathcal{L}_{\text{pose}}=\frac{1}{N_J}\sum_j^{N_J}\Big(\abs{u_j^{\text{GT}}-\hat{u}_j} + \abs{v_j^{\text{GT}}-\hat{v}_j}\Big).
\end{align}
To deal with multiple persons in the image, we focus on a target person by zeroing out the regions of the heatmap outside the bounding box of that person from the object detector.

\subsubsection{Depth Estimation Branch.}
After we obtain the heatmaps, we use them as attention masks to guide the network into focusing on specific regions of the image related to the target person. More specifically, we only care about features from pixel locations that are close to the joints of the target person. The intuition behind our design choice is that joint locations contain more scale-related information than the larger yet less discriminative areas such as the whole upper body trunk. Attention-guided feature pooling is also adopted in other tasks such as action recognition \cite{luvizon20182d} and hand pose estimation \cite{iqbal2018hand}.
We compute the weighted sum feature vector $\mathbf{d}$ for each joint $j$ 
from the feature block $\mathbf{F}_\text{depth}$ as:
\begin{align}
    \mathbf{d}^{(j)}=\sum_{(u,v)=(0,0)}^{(W-1,H-1)}\hat{\mathbf{H}}_{u,v}^{(j)}\cdot {\mathbf{F}_{\text{depth}}}_{u,v}.
\end{align}
To effectively aggregate features corresponding to different joint types, we formulate a standard Graph Neural Network (GNN) where each node represents one joint type, \textit{e.g.}, elbow, knee, \textit{etc}. The aggregated features $\mathbf{d}^{(j)}$ for each joint type $j$ is fed into the corresponding node $X_{in}^{(j)}$ in the graph as input. Each layer of the GNN is defined as:
\begin{align}
    X_{out}^{(i)}=\sigma\Big(\Tilde{a}_{ii}f_{self}(X_{in}^{(i)};\Theta_{self}) + \sum_{j\ne i}\Tilde{a}_{ij}f_{inter}(X_{in}^{(j)};\Theta_{inter})\Big).
\label{eqn:gnn}
\end{align}
The feature of each input node $X_{in}$ undergoes the linear mappings $f_{self}(.)$ and $f_{inter}(.)$ that are parametrized by $\Theta_{self}$ and $\Theta_{inter}$, respectively. The output of the node, i.e. $X_{out}^{(i)}$ is computed from a weighted aggregation of
$f_{self}(.)$ and $f_{inter}(.)$ of all other nodes.
The weighting factor $\Tilde{a}_{ij}$ is an element of the normalized adjacency matrix $\Tilde{A} \in \mathbb{R}^{N_j\times N_j}$ that controls the extent of influence of the nodes on each other.
The original adjacency matrix is $A \in \{0,1\}^{N_j\times N_j}$; an element $a_{ij}$ equals 1 if there is a skeletal link between joint $i$ and $j$, \textit{e.g.} left knee to left ankle, or otherwise 0.
$\Tilde{A} \in \mathbb{R}^{N_j\times N_j}$ is obtained by applying $L1$-normalization on each row of $A$.
The non-linearity function $\sigma(.)$ is implemented with a Batch Normalization followed by a ReLU.
We stack $L$ GNN layers in total.
After the last GNN layer, we merge the feature output from each node with an average pooling operation.

\subsubsection{Target output formulation}
Inspired by the work \cite{fu2018deep} for scene depth estimation,
we formulate the depth estimation as a classification problem
instead of directly regressing the numerical value of depth.
We follow the practice in \cite{fu2018deep} to discretize the log-depth space into a preset number of bins, $N_\mathbf{B}$. We compute:
\begin{align}
    b(d)=\frac{\log{d}-\log{\alpha}}{{\log{\beta}-\log{\alpha}}}\cdot (N_\mathbf{B}-1),
\end{align}
where $\lfloor b \rceil$ gives the bin index of the depth, and the depth $d$ of a pose is assumed to be within the range $[\alpha, \beta]$. Here $\lfloor . \rceil$ is the round-off to the nearest integer operator. 
To eliminate quantization errors, we assign non-zero values to two consecutive bins $i$ and $i+1$, where $i \leq b < i+1$. This operation is similar to the weights in bi-linear interpolation. For example, the ground truth values of the bins are given by $\mathbf{B} = [0, 0, 0.6, 0.4, 0]$ for $N_\mathbf{B}=5$ and $b=2.4$.
Consequently, $\mathbf{B}$ is a 1D heatmap that can achieve any level of precision with a sufficiently accurate categorical estimation on the bins.

Since a different focal length of the camera affects the scale of a target person in the image, it is unrealistic to estimate the absolute depth from images taken by any arbitrary camera. To alleviate this problem, we normalize out the camera intrinsic parameters by replacing the target $d$ with $\hat{d}=d/f$, where $f$ is the focal length of camera. We approximate with $\hat{d}=d/\sqrt{f_x\cdot f_y}$ in our experiments since the focal lengths in $x$ and $y$ directions are usually very close.
Finally, we add a fully connected layer after the pooled feature from the last GNN layer to regress the $N_\mathbf{B}$ values of the bins $\hat{\mathbf{B}}$. Softmax operation is used to normalize the output into a valid distribution. We transform $\hat{\mathbf{B}}$ back to the estimated depth $d$ of the root joint by:
\begin{equation}
    d=\exp{\Big[\frac{\hat{b}}{N_\mathbf{B}-1}\cdot (\log{\beta}-\log{\alpha})+\log{\alpha}\Big]}\cdot \sqrt{f_x \cdot f_y},~\text{where}~ \hat{b}=\sum_{i=0}^{N_\mathbf{B}-1}\hat{\mathbf{B}}_i \cdot i.
\end{equation}
Similar to the soft-argmax operation used to transform heatmaps to joint locations, $\hat{b}$ is the weighted sum of the bin indices with the estimated heatmap $\hat{\mathbf{B}}$.
To supervise the learning of the depth estimation branch, we adopt cross-entropy loss on the estimated bins $\hat{\mathbf{B}}$ and $L1$ loss on $\hat{b}$ as follows:
\begin{equation}
    \mathcal{L}_{\text{bins}}=-\sum_{i=0}^{N_\mathbf{B}-1}\mathbf{B}_i^{\text{GT}} \cdot \log{\hat{\mathbf{B}}_i},~\text{and}~~ \mathcal{L}_{\text{idx}}=\abs{b^{\text{GT}}-\hat{b}}.
\end{equation}
We train the whole framework with losses from the pose estimation and depth estimation branches:
\begin{equation}
    \mathcal{L}=\lambda_{\text{hm}}\mathcal{L}_{\text{hm}}+\lambda_{\text{pose}}\mathcal{L}_{\text{pose}}+\lambda_{\text{bins}}\mathcal{L}_{\text{bins}}+\lambda_{\text{idx}}\mathcal{L}_{\text{idx}}.
\label{eqn:loss}
\end{equation}

\section{Experiments}

\subsection{Datasets and Evaluation Metrics}

\subsubsection{Human3.6M dataset.}
Human3.6M dataset \cite{ionescu2014human3} is currently the largest publicly available dataset for human 3D pose estimation. The dataset consists of 3.6 million video frames captured by MoCap system in a constrained indoor studio environment. 11 actors performing 15 activities are captured from 4 camera viewpoints. 3D ground truth poses in world coordinate system and camera extrinsic (rotation and translation with respect to world coordinate) and intrinsic parameters (focal length and principal point) are available. We follow previous works that five subjects (S1, S5, S6, S7, S8) are used in training and two subjects (S9 and S11) are used for evaluation. We use every 5th and 64th frames in each video for training and evaluation respectively. No extra 2D pose dataset is used to augment the training. We follow the metric Mean Root Position Error (MRPE) proposed in \cite{Moon_2019_ICCV_3DMPPE}
to evaluate the root localization accuracy. 
Specifically, we consider the Euclidean distance between the estimated and the ground truth 3D coordinate of the root joint.

\subsubsection{MuCo-3DHP and MuPoTS-3D datasets.}
MuCo-3DHP and MuPoTS-3D are two datasets proposed by Mehta \textit{et al.} \cite{singleshotmultiperson2018} to evaluate multi-person 3D pose estimation performance. The training set MuCo-3DHP is a composite dataset which merges randomly sampled 3D poses from single-person 3D human pose dataset MPI-INF-3DHP \cite{mono-3dhp2017} to form realistic multi-person scenes. The test set MuPoTS-3D is a markerless motion captured multi-person dataset including both indoor and outdoor scenes. We use the same set of MuCo-3DHP synthesized images from \cite{Moon_2019_ICCV_3DMPPE} for a fair comparison. No extra 2D pose dataset is used to augment the training.
For evaluation of multi-person root joint localization, we follow \cite{Moon_2019_ICCV_3DMPPE} to report the average precision and recall of 3D root joint location under different thresholds. A root joint with a smaller distance to the matched ground truth root joint location than a threshold is considered a true positive estimation.
We follow \cite{Moon_2019_ICCV_3DMPPE} to report $\text{3DPCK}_{abs}$
for evaluation of the root-aware 3D pose estimation,
where 3DPCK (3D percentage of correct keypoints) for the estimated poses is evaluated without root alignment.
3DPCK treats an estimated joint as correct if it is within 15 cm distance from the matched ground truth joint.
Although our framework does not focus on root-relative 3D pose estimation, we also report the root-aligned $\text{3DPCK}_{rel}$ to show that accurate root localization also benefits the precision of 3D pose estimation.

\subsection{Implementation Details}
\label{sec:implementation_details}
Following previous work \cite{Moon_2019_ICCV_3DMPPE}, we use Mask R-CNN \cite{he2017mask} as our person detector due to its high performance and generalizability to in-the-wild images.
For single-person 3D pose estimation, we use the volumetric-based 3D pose estimator by \cite{sun2018integral}.
Instead of cropping out areas of interest using bounding boxes, we keep the original scale of image and crop out a fixed size patch centered around the bounding box, or the principal point if no bounding box is provided in the single-person scenario. The cropped out image is then rescaled to $256\times 256$ and used as input to our network. The output resolution of the heatmap is $64\times 64$.
We use 2 layers of GNN operations in the depth estimation branch.
We set the standard deviation of the Gaussian peak in the ground truth heatmap to be 0.75, and 
the bin range of $d/f$ to $[\alpha =1.0, \beta =8.0]$ for a reasonably sufficient range of the depth.
We do not see much performance change when different number of bins $N_\mathbf{B}$ are used. All results of the experiments shown in the paper are obtained with $N_\mathbf{B} = 71$.
We set $\lambda$ in Eq. \ref{eqn:loss} to balance the four loss terms to same order of magnitudes.
For training, we use Adam optimizer \cite{kingma2014adam} with learning rate 1e-4 and batch size 16. We train the model for 200k steps and decay the learning rate with a factor of 0.8 at every 20k steps.
The evaluation of each image takes around 7ms with our root joint localization HDNet.

\subsection{Results on Human3.6M}

\setlength{\tabcolsep}{4pt}
\begin{table}[t]
\begin{center}
\caption{MRPE results comparison with state-of-the-arts on the Human3.6M dataset. $\text{MRPE}_x$, $\text{MRPE}_y$, and $\text{MRPE}_z$ are the average errors in $x$, $y$, and $z$ axes, respectively.}
\label{table:mrpe_h36m}
\begin{tabular}{l|llll}
\hline\noalign{\smallskip}
Method & MRPE & $\text{MRPE}_x$ & $\text{MRPE}_y$ & $\text{MRPE}_z$\\
\noalign{\smallskip}
\hline
\noalign{\smallskip}
Baseline & 267.8 & 27.5 & 28.3 & 261.9\\
Baseline w/o limb joints & 226.2 & 24.5 & 24.9 & 220.2\\
Baseline with RANSAC & 213.1 & 24.3 & 24.3 & 207.1\\
RootNet \cite{Moon_2019_ICCV_3DMPPE} & 120.0 & 23.3 & 23.0 & 108.1\\
\noalign{\smallskip}
\hline
\noalign{\smallskip}
\textbf{Ours} & \textbf{77.6} & \textbf{15.6} & \textbf{13.6} & \textbf{69.9}\\
\noalign{\smallskip}
\hline
\end{tabular}
\end{center}
\end{table}
\setlength{\tabcolsep}{1.4pt}

The root joint localization results on Human3.6M dataset are shown in Table \ref{table:mrpe_h36m}. The baselines reported in the top 3 rows follow a two-stage approach, where 2D pose \cite{sun2018integral} and 3D pose \cite{martinez2017simple} are estimated separately, and an optimization process is adopted to obtain the global root joint location that minimizes the reprojection error. ``w/o limb joints" refers to optimization using only head and body trunk joints. ``with RANSAC" refers to randomly sampling the set of joints used for optimization with RANSAC.
The baseline results are taken from the figures reported in \cite{Moon_2019_ICCV_3DMPPE}.
We also compare with the state-of-the-art approach \cite{Moon_2019_ICCV_3DMPPE}.
It can be seen from Table \ref{table:mrpe_h36m} that optimization-based methods can achieve reasonable results, but with limited accuracy due to the errors from both the 2D and 3D estimation stages.
Our root joint localization framework achieves 69.9mm depth estimation error in $\text{MRPE}_z$ with a 35\% improvement over \cite{Moon_2019_ICCV_3DMPPE}. Since our approach uses the original scale image without scaling to person bounding box size which limits the 2D $(u,v)$ localization precision, we also adopt a state-of-the-art 2D pose estimator CPN \cite{chen2018cascaded} within the person bounding box area to further refine the $(u,v)$ localization. Our MRPE for root joint achieves an overall performance of 77.6mm which significantly outperforms the state-of-the-art.

\setlength{\tabcolsep}{4pt}
\begin{table}[t]
\begin{center}
\caption{Root joint localization accuracy comparison in average precision and recall with state-of-the-arts on MuPoTS-3D dataset.}
\label{table:ap_mupo}
\begin{tabular}{l|llll|llll}
\hline\noalign{\smallskip}
Method & $\text{AP}_{25}^{\text{root}}$ & $\text{AP}_{20}^{\text{root}}$ & $\text{AP}_{15}^{\text{root}}$ & $\text{AP}_{10}^{\text{root}}$ & $\text{AR}_{25}^{\text{root}}$ & $\text{AR}_{20}^{\text{root}}$ & $\text{AR}_{15}^{\text{root}}$ & $\text{AR}_{10}^{\text{root}}$\\
\noalign{\smallskip}
\hline
\noalign{\smallskip}
RootNet \cite{Moon_2019_ICCV_3DMPPE} & 31.0 & 21.5 & 10.2 & 2.3 & 55.2 & 45.3 & 31.4 & 15.2\\
\noalign{\smallskip}
\hline
\noalign{\smallskip}
\textbf{Ours} & \textbf{39.4} & \textbf{28.0} & \textbf{14.6} & \textbf{4.1} & \textbf{59.8} & \textbf{50.0} & \textbf{35.9} & \textbf{19.1}\\
\noalign{\smallskip}
\hline
\end{tabular}
\end{center}
\end{table}
\setlength{\tabcolsep}{1.4pt}

\setlength{\tabcolsep}{4pt}
\begin{table}[t]
\begin{center}
\caption{Sequence-wise $\text{3DPCK}_{abs}$ comparison with state-of-the-arts on MuPoTS-3D dataset. Accuracy is measured on matched ground-truths.}
\label{table:seq_pck_abs_mupo}
\resizebox{\textwidth}{!}{
\begin{tabular}{l|cccccccccc|c}
\hline\noalign{\smallskip}
Method & S1 & S2 & S3 & S4 & S5 & S6 & S7 & S8 & S9 & S10 & -\\
\noalign{\smallskip}
\hline
\noalign{\smallskip}
RootNet \cite{Moon_2019_ICCV_3DMPPE} & \textbf{59.5} & \textbf{45.3} & 51.4 & \textbf{46.2} & \textbf{53.0} & \textbf{27.4} & 23.7 & 26.4 & 39.1 & 23.6 & -\\
\textbf{Ours} & 21.4 & 22.7 & \textbf{58.3} & 27.5 & 37.3 & 12.2 & \textbf{49.2} & \textbf{40.8} & \textbf{53.1} & \textbf{43.9} & -\\
\noalign{\smallskip}
\hline
\noalign{\smallskip}
Method & S11 & S12 & S13 & S14 & S15 & S16 & S17 & S18 & S19 & S20 & Avg\\
\noalign{\smallskip}
\hline
\noalign{\smallskip}
RootNet \cite{Moon_2019_ICCV_3DMPPE}  & 18.3 & 14.9 & 38.2 & \textbf{29.5} & 36.8 & 23.6 & 14.4 & 20.0 & 18.8 & 25.4 & 31.8\\
\textbf{Ours} & \textbf{43.2} & \textbf{43.6} & \textbf{39.7} & 28.3 & \textbf{49.5} & \textbf{23.8} & \textbf{18.0} & \textbf{26.9} & \textbf{25.0} & \textbf{38.8} & \textbf{35.2}\\
\noalign{\smallskip}
\hline
\end{tabular}
}
\end{center}
\end{table}
\setlength{\tabcolsep}{1.4pt}

\setlength{\tabcolsep}{4pt}
\begin{table}[t]
\begin{center}
\caption{Joint-wise $\text{3DPCK}_{abs}$ comparison with state-of-the-arts on MuPoTS-3D dataset. Accuracy is measured on matched ground-truths.}
\label{table:joint_pck_abs_mupo}
\begin{tabular}{l|cccccccc|c}
\hline\noalign{\smallskip}
Method & Head & Neck & Shoulder & Elbow & Wrist & Hip & Knee & Ankle & Avg\\
\noalign{\smallskip}
\hline
\noalign{\smallskip}
RootNet \cite{Moon_2019_ICCV_3DMPPE} & 37.6 & 35.6 & 34.0 & 34.1 & 30.7 & 30.6 & 31.3 & 25.3 & 31.8\\
\textbf{Ours} & \textbf{38.3} & \textbf{37.8} & \textbf{36.2} & \textbf{37.4} & \textbf{34.0} & \textbf{34.9} & \textbf{36.4} & \textbf{29.2} & \textbf{35.2}\\
\noalign{\smallskip}
\hline
\end{tabular}
\end{center}
\end{table}
\setlength{\tabcolsep}{1.4pt}

\subsection{Results on MuPoTS-3D}

\subsubsection{Root Joint Localization.}
To evaluate our root joint localization performance on the multi-person MuPoTS-3D dataset, we estimate the root joint 3D coordinate for each bounding box detected from the object detector. All root joint candidates are matched with the ground truth root joints, and only candidates with distance to the matched ground truth lesser than a threshold are considered as an accurate estimate. We then analyze the average precision and recall over the whole dataset under various settings of thresholds ranging from 25cm to 10cm. The results are shown in Table \ref{table:ap_mupo}. Our method achieves much higher AP and AR consistently across all levels of thresholds compared to the state-of-the-art approach \cite{Moon_2019_ICCV_3DMPPE}.

\setlength{\tabcolsep}{4pt}
\begin{table}[t]
\begin{center}
\caption{Sequence-wise $\text{3DPCK}_{rel}$ comparison with state-of-the-arts on MuPoTS-3D dataset. Accuracy is measured on matched ground-truths.}
\label{table:seq_pck_rel_mupo}
\resizebox{\textwidth}{!}{
\begin{tabular}{l|cccccccccc|c}
\hline\noalign{\smallskip}
Method & S1 & S2 & S3 & S4 & S5 & S6 & S7 & S8 & S9 & S10 & -\\
\noalign{\smallskip}
\hline
\noalign{\smallskip}
Rogez \textit{et al.} \cite{rogez2017lcr} & 69.1 & 67.3 & 54.6 & 61.7 & 74.5 & 25.2 & 48.4 & 63.3 & 69.0 & 78.1 & -\\
Mehta \textit{et al.} \cite{singleshotmultiperson2018} & 81.0 & 65.3 & 64.6 & 63.9 & 75.0 & 30.3 & 65.1 & 61.1 & 64.1 & 83.9 & -\\
Rogez \textit{et al.} \cite{rogez2019lcr} & 88.0 & 73.3 & 67.9 & 74.6 & 81.8 & 50.1 & 60.6 & 60.8 & 78.2 & 89.5 & -\\
RootNet \cite{Moon_2019_ICCV_3DMPPE} & \textbf{94.4} & 78.6 & 79.0 & 82.1 & 86.6 & 72.8 & \textbf{81.9} & 75.8 & \textbf{90.2} & 90.4 & -\\
\textbf{Ours} & \textbf{94.4} & \textbf{79.6} & \textbf{79.2} & \textbf{82.4} & \textbf{86.7} & \textbf{73.0} & 81.6 & \textbf{76.3} & 90.1 & \textbf{90.5} & -\\
\noalign{\smallskip}
\hline
\noalign{\smallskip}
Method & S11 & S12 & S13 & S14 & S15 & S16 & S17 & S18 & S19 & S20 & Avg\\
\noalign{\smallskip}
\hline
\noalign{\smallskip}
Rogez \textit{et al.} \cite{rogez2017lcr} & 53.8 & 52.2 & 60.5 & 60.9 & 59.1 & 70.5 & 76.0 & 70.0 & 77.1 & 81.4 & 62.4\\
Mehta \textit{et al.} \cite{singleshotmultiperson2018} & 72.4 & 69.9 & 71.0 & 72.9 & 71.3 & 83.6 & 79.6 & 73.5 & 78.9 & \textbf{90.9} & 70.8\\
Rogez \textit{et al.} \cite{rogez2019lcr} & 70.8 & 74.4 & 72.8 & 64.5 & 74.2 & 84.9 & 85.2 & 78.4 & 75.8 & 74.4 & 74.0\\
RootNet \cite{Moon_2019_ICCV_3DMPPE} & \textbf{79.4} & \textbf{79.9} & 75.3 & 81.0 & \textbf{81.1} & 90.7 & \textbf{89.6} & 83.1 & 81.7 & 77.3 & 82.5\\
\textbf{Ours} & 77.9 & 79.2 & \textbf{78.3} & \textbf{85.5} & \textbf{81.1} & \textbf{91.0} & 88.5 & \textbf{85.1} & \textbf{83.4} & 90.5 & \textbf{83.7}\\
\noalign{\smallskip}
\hline
\end{tabular}
}
\end{center}
\end{table}
\setlength{\tabcolsep}{1.4pt}

\subsubsection{Camera-space absolute 3D pose estimation.}
We also evaluate the camera-space absolute 3D pose estimation performance with $\text{3DPCK}_{abs}$.
$\text{3DPCK}_{abs}$ compares the estimated 3D pose with the matched ground truth pose in the camera coordinate space without root alignment, thus requires highly accurate root joint localization. We use the same 3D pose estimator \cite{sun2018integral} as the state-of-the-art root joint localization method \cite{Moon_2019_ICCV_3DMPPE} for a fair comparison. Results in Table \ref{table:seq_pck_abs_mupo} show that our method consistently outperforms the state-of-the-art in most of the test sequences and achieves a 35.2\% average 3DPCK (3.4\% improvement). The performance breakdown of all joint types is shown in Table \ref{table:joint_pck_abs_mupo}.

\subsubsection{Root-relative 3D pose estimation.}
The state-of-the-art root-relative 3D pose estimator \cite{sun2018integral} adopts a volumetric output representation and estimates the root-relative depth for each joint. 
Absolute root joint depth has to be available
to recover the 3D pose through back-projection. We follow \cite{Moon_2019_ICCV_3DMPPE} and use our estimated root depth to back-project the 3D pose and evaluate the root-relative 3D pose estimation accuracy with $\text{3DPCK}_{rel}$ after root joint alignment. Results are shown in Table \ref{table:seq_pck_rel_mupo}, where our method outperforms the previous best performance by 1.2\% average 3DPCK. This demonstrates that more accurate root localization also benefits the precise 3D pose estimation in volumetric-based approaches \cite{pavlakos2017coarse,sun2018integral,Moon_2019_ICCV_3DMPPE}.

\subsection{Ablation Studies}

\setlength{\tabcolsep}{4pt}
\begin{table}[t]
\begin{center}
\caption{Ablation studies on components of the framework. Depth error $\text{MRPE}_z$ (mm) on Human3.6M dataset and $\text{AP}_{25}^{\text{root}}$ (\%) on MuPoTS-3D dataset are measured.}
\label{table:ablation}
\begin{tabular}{l|c|c}
\hline\noalign{\smallskip}
Method & $\text{MRPE}_z(\downarrow)$ & $\text{AP}_{25}^{\text{root}}(\uparrow)$\\
\noalign{\smallskip}
\hline
\noalign{\smallskip}
RootNet \cite{Moon_2019_ICCV_3DMPPE} & 108.1 & 31.0\\
Ours direct regression & 94.5 & 27.3\\
Ours shared feature branch & 72.0 & 31.9\\
Ours w/o GNN & 72.9 & 32.7\\
Ours w/o HM pooling & 71.8 & 26.0\\
\textbf{Ours (full)} & \textbf{69.9} & \textbf{39.4}\\
\noalign{\smallskip}
\hline
\end{tabular}
\end{center}
\end{table}
\setlength{\tabcolsep}{1.4pt}

We conduct ablation studies to show how each component in our framework affects the root joint localization accuracy. We evaluate the depth estimation accuracy $\text{MRPE}_z$ on Human3.6M dataset and the root joint localization $\text{AP}_{25}^{\text{root}}$ on MuPoTS-3D dataset with different variants of our framework in Table \ref{table:ablation}.
The state-of-the-art approach \cite{Moon_2019_ICCV_3DMPPE} is also included for comparison.
\begin{itemize}
\item ``Ours direct regression":
Performance drop (by 24.6mm and 12.1\%) with directly regressing target depth instead of performing classification over binning shows the effectiveness of formulating the depth estimation as a classification task.
\item ``Ours shared feature branch": One single multi-scale feature branch is kept after FPN, which means $\mathbf{F}_\text{pose}$ and $\mathbf{F}_\text{depth}$ use the same feature representation. This setting causes performance to drop (by 2.1mm and 7.5\%), and thus demonstrates that the features used for pose estimation and depth estimation are not highly correlated.
\item ``Ours w/o GNN": We replace the GNN layers in our depth estimation branch with same number of fully-connected layers and observe a performance drop (by 3mm and 6.7\%), showing the effectiveness of the graph neural network in propagating and refining the features extracted for different types of joints.
\item ``Ours w/o HM pooling": We remove feature pooling with estimated heatmaps as mask in the depth estimation branch and instead apply a global average pooling to obtain a single feature vector. The GNN layers are replaced with fully-connected layers since we do not explicitly differentiate between different joint types. We observe a performance drop (by 1.9mm and 13.4\%), which demonstrates the effectiveness of utilizing estimated pose as attention mask for useful feature aggregation.
\end{itemize}

\subsection{Discussions}

\begin{figure}[t]
\centering
\includegraphics[width=0.5\textwidth,trim={10 32 10 10},clip]{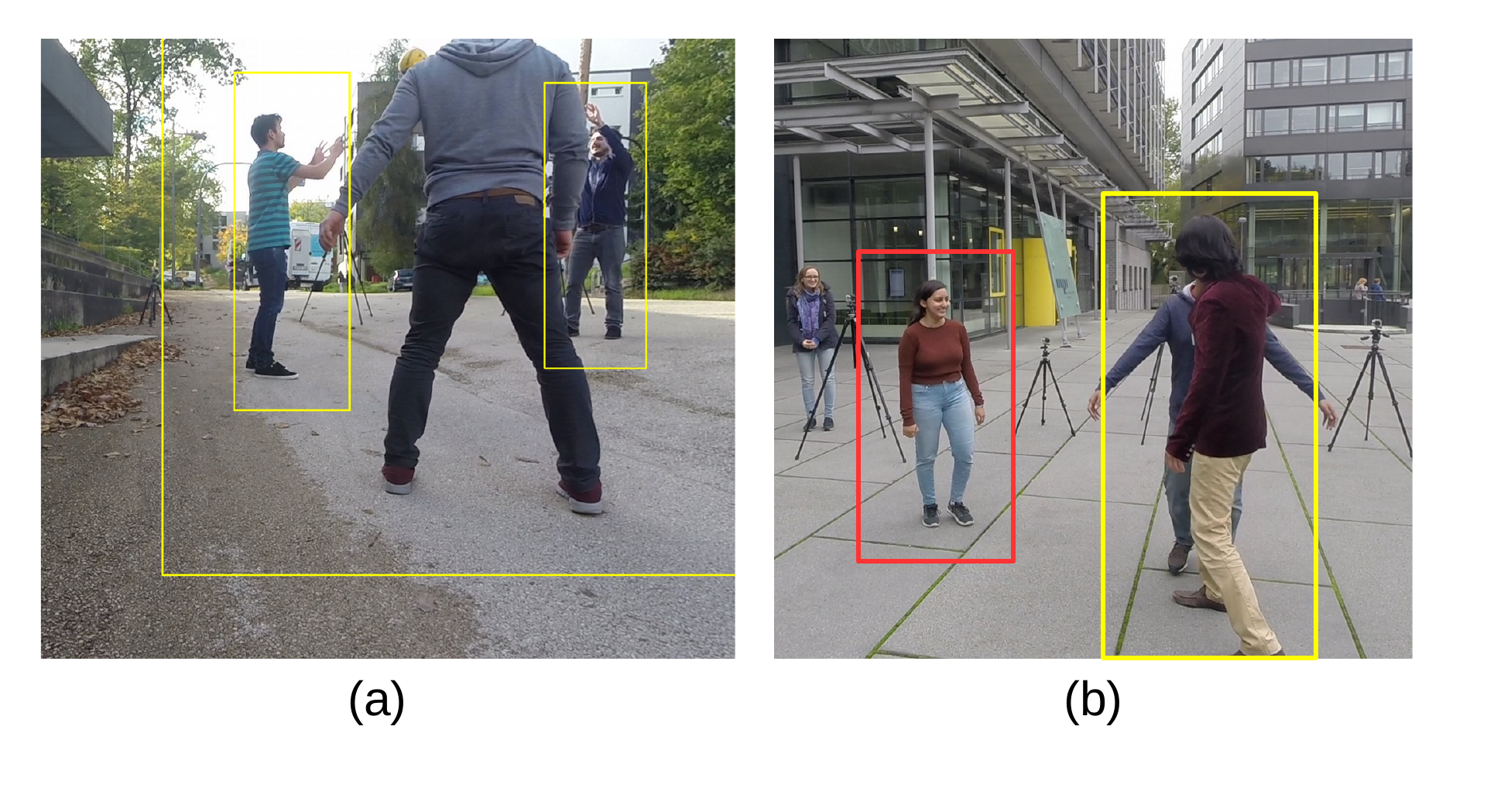}
\caption{Typical errors in multi-person root localization. (a) Close and overlapping bounding box regions. (b) Different sizes of target persons.}
\label{fig:error}
\end{figure}

We analyze the root joint localization results on the challenging multi-person dataset MuPoTS-3D and observe several sources of large errors as shown in Figure \ref{fig:error}:
(1) Bounding boxes for two persons tend to have overlapping areas when 
the person closer to the camera partially occludes the other person farther away (Figure \ref{fig:error}(a)). Masking the heatmaps with bounding box cannot effectively remove undesired regions of information and consequently the depth estimation for both persons are affected. The problem of fine-grained target person segmentation will be of interest for future research.
(2) Since monocular depth estimation relies on prior knowledge such as typical scale of human bodies, estimation tends to be erroneous when the size of target person is far away from the ``average" size, \textit{e.g.}, the target is a child or a relatively short person (Figure \ref{fig:error}(b)). Research on person 3D size estimation may complement our depth estimation task and improve the generalizability to persons of different sizes.

\begin{figure}[t]
\centering
\includegraphics[width=0.85\textwidth,trim={10 10 10 10},clip]{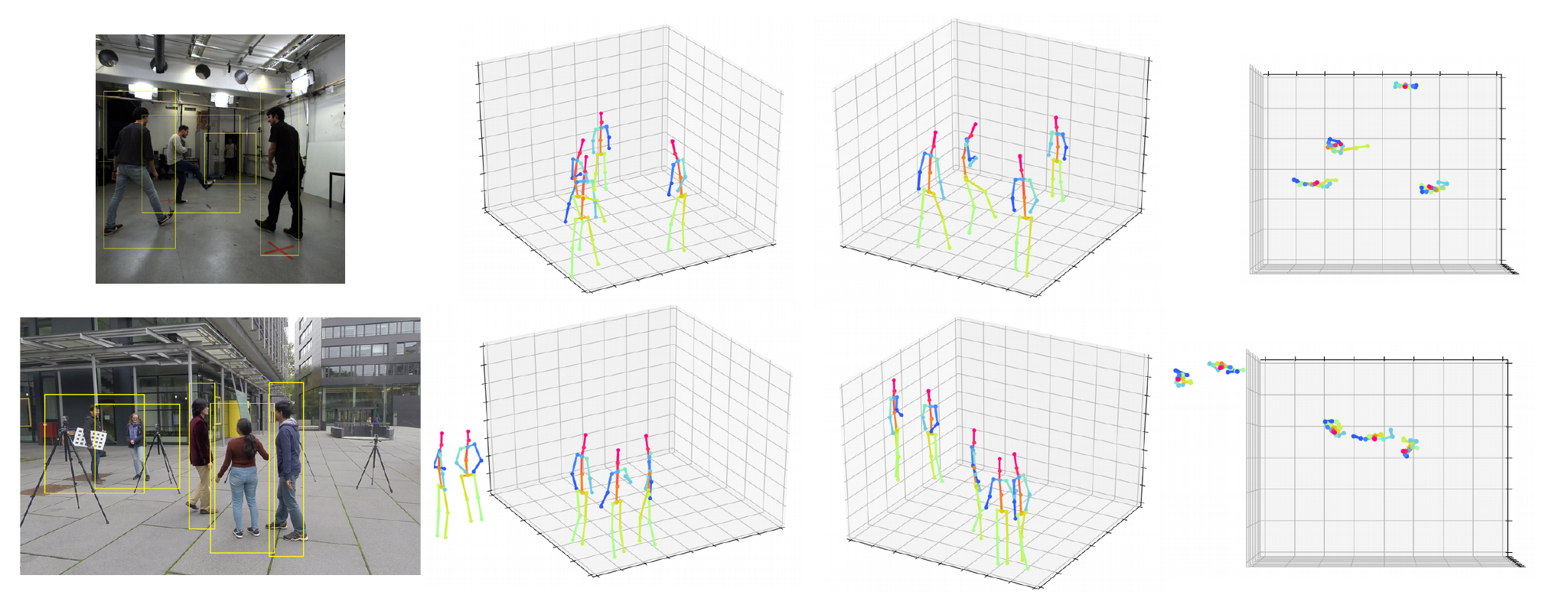}
\caption{Qualitative results on MuPoTS-3D dataset. Columns are: (1) image with bounding boxes (2) left-front view (3) right-front view (4) top-down view}
\label{fig:qualitative}
\end{figure}

\section{Conclusions}

In this work, we proposed the Human Depth Estimation Network (HDNet), an end-to-end framework to address the problem of accurate root joint localization for multi-person 3D absolute pose estimation. Our HDNet utilizes deep features and demonstrates the capability to precisely estimate depth of root joints. We designed a human-specific pose-based feature aggregation process in the HDNet to effectively pool features from regions of human body joints. Experimental results on multiple datasets showed that our framework significantly outperforms the state-of-the-art in both root joint localization and 3D pose estimation.

\clearpage
%
%
\bibliographystyle{splncs04}
\bibliography{hdnet}
\end{document}